%% file: main.tex
\documentclass[11pt]{article}
\usepackage[final]{acl}
\usepackage{graphicx} % Required for inserting images
 
\usepackage{times}
\usepackage{latexsym}
\usepackage[T1]{fontenc}
\usepackage[utf8]{inputenc}
\usepackage{microtype}
\usepackage{inconsolata}
\usepackage{booktabs}
\usepackage{paralist}
\usepackage{multirow}
\usepackage{pdfpages}
\usepackage{hyperref}

\usepackage{annotates}

\title{Comparing and Modeling Argumentation in \\ German Political Communication across Arenas}
\author{Nina Vikhrova$^1$ and Johannes Kühling$^2$ and Sebastian Haunss$^2$ and Sebastian Padó$^1$ \\
$^1$IMS, Universität Stuttgart,
$^2$SOCIUM, Universität Bremen \\ 
 \texttt{\{nina.vikhrova|pado\}@ims.uni-stuttgart.de}, 
 \texttt{\{kuehling|haunss\}@uni-bremen.de}}
 
\date{March 2026}

\begin{document}

\maketitle

\begin{abstract}
Deliberation, involving the formulation and exchange of arguments, forms an integral part of political decision making in democracies. Argumentation patterns however differ substantially across different political arenas, such as plenary speeches and committee meetings. However, despite a lot of interest in argumentation, there is comparatively little computational work on analyzing differences in patterns of political argumentation between arenas. \\
Our work addresses this research gap. First, we present a 17k-sentence corpus with annotation for argumentative passages (argument and their justifications, both their boundaries and their categories) across three German political arenas (plenary speeches, committee meetings, and press conferences), keeping the topic (COVID-19) constant. Our analysis of the corpus finds that contrary to expectations, justification by domain-specific expertise is more frequent in press conferences than in committee meetings. Second, we present a pilot study on automatically identifying such argumentative passages. The results show that boundaries are hard to pin down, and models predictions additionally suffer from confirmation bias.
\end{abstract}

\section{Introduction}
\label{sec:intro}
% \begin{itemize}
% \item Argumentation structure (argumentative + justificational passages) in political discourse
% \item Link to 'context matters' conference theme (https://konvens2026.uni-hamburg.de/index.php/first-call-for-conference-papers/)
% \item One important legitimation: expertise
% \item To what extent is this actually the case? What does actual argumentation look like?
% \item Our investigation: on the domain of medical discussion: COVID
% \item Contributions:
%     \begin{itemize}
%         \item Dataset
%         \item Baseline modeling
%     \end{itemize}
%   \end{itemize}
  
Even if existing democracies are often far from fulfilling the ideal of deliberative democracies \cite{Baechtiger_Dryzek_Mansbridge_Warren_2018}, political decision-making always contains elements of deliberation in parliaments, committees, and other arenas. In these fora, actors make claims and argue to support their positions and to attack opposing political actors. To advance their goal, they use various rhetorical and argumentation strategies, one of which is what Reyes has called ``voices of expertise'' \cite[786]{Reyes_2011}, that is, the referral to experts or to expert knowledge in order to legitimise one’s claim.

One should expect that argumentation strategies, in particular with regard to the appeal to expertise, are context dependent: Argumentation in plenary speeches should differ from argumentation in committees because the former address a general public, while the latter address mainly other MPs who are likely experts or at least experienced in the respective field. Consequently, one could expect that expert argumentation should be more common in committees than in plenary speeches. But studies so far have generally been limited to specific debates in one specific forum at a time. Comparative studies on argumentation strategies are very rare \cite[e.g.][]{Karlsson_Persson_Martensson_2024} because of the required amount of annotation. 
%An exception is \citet{poiaganova-stede-2025-debates} who consider arguments in US/UN political speeches.

In our study, we aim to make one step forward in the study of political debate across arenas by annotating, evaluating, and modeling arguments from three different corpora of political speech in Germany: plenary speeches, health committee meetings, and press conferences. We focus our analysis on the political discourse in these three arenas during the COVID pandemic in Germany (2020--2022). We choose this specific period and topic since during the COVID pandemic the reference to (medical) expert knowledge was omnipresent and we can therefore expect to see the use of expert arguments by many political actors, independent of whether they are themselves medical experts or not. Focusing on one domain also reduces the impact of topic shifts on our analysis.

Our paper makes two main contributions:
\begin{enumerate}
\item We present and analyse
a manually annotated dataset of political discourse from  three different German federal-level political arenas -- to our knowledge, the first such corpus. It makes explicit argumentative passages, their justifications, and the type of justification used. Contrary to expectations, expert arguments play a bigger role in press conferences than in committee meetings.

\item We present a pilot study on the ability of  LLMs to identify argumentative passages in political arenas within and across fora. This would be a first step towards
reducing manual effort for such analyses. We find, however, that argumentative passages are hard to identify: Many models are reluctant to conclude that documents are non-argumentative.
\end{enumerate}

\section{Related Work}

%\begin{itemize}
%    \item political science background (SH)
%    \item argument mining (SP)
%\end{itemize}

\paragraph{Argument Mining for Political Texts.} In computational linguistics, the automatic extraction of arguments (``argument mining'') has grown rapidly over the last ten years, but has focused mainly on text types where ample data was available, including social media and student texts, see  \citet{cabrio2018five} for a review. Generalizing argument mining methods across text types is often challenging \citep{daxenberger-etal-2017-essence,schaefer-etal-2022-selecting}. When it comes to political texts, most studies focus on uniform datasets of current interest, typically election campaigns \citep{Lippi_Torroni_2016,Visser2020Argumentation}.
An exception is \citet{poiaganova-stede-2025-debates} who evaluate various argument mining-related tasks within and across two political arenas (US presidential debates and UN security council speeches) and find reasonable generalization for most tasks across the two arenas. Although argument mining in parliamentary debates is already an established field of research, our approach is timely because it applies a zero-shot large language model to the task. Moreover, our annotation scheme extends existing political argumentation corpora by distinguishing different types of justification, including domain-specific justifications tailored to the analysis of expert argumentation.

\paragraph{Political Argument Quality.}
From a political science perspective, Steenbergen et al.’s Discourse Quality Index (DCI) is a normative evalutation of argumentation and in essence attempts to measure to which degree a debate fulfils the Habermasian ideal of deliberative politics \cite{Steenbergen_Baechtiger_Spoerndli_Steiner_2003, Steffensmeier_Schenck-Hamlin_2008}. Research from a critical discourse analysis perspective puts a stronger focus on the use of various legitimization strategies, namely legitimization through emotions, through references to the future, through rationality, expertise, and altruism \cite{Reyes_2011}. In the context of argument mining, it has been found that argument quality is best considered as a multidimensional phenomenon. \citet{Wachsmuth_Naderi_Hou_Bilu_Prabhakaran_Thijm_Hirst_Stein_2017} have proposed a taxonomy for argumentation quality based on the cogency, reasonableness, and effectiveness of an argument.

\paragraph{Political Arguments across Arenas.}
Regarding arena specific argumentation strategies, political science research has highlighted differences between ``debating'' (e.g. UK) and ``working'' (e.g. Denmark, Sweden) parliaments, and between parliamentary ``frontstage'' and ``committee backstage'' \cite{Karlsson_Persson_Martensson_2024} and pointed to more pragmatic argumentation in committees \cite{Andone_2016}. On the computational side, there are few studies that compare properties of arguments across arenas. \citet{Reinig_Rehbein_Ponzetto_2024} characterize political speech in terms of speech acts, which are however distributed fairly complementarily to arguments.  As mentioned above, \citet{poiaganova-stede-2025-debates} consider two political arenas. Their mostly good generalization results imply that the arenas contain arguments with similar properties, but a deeper comparative analysis (e.g., with regard to expertise) was beyond the scope of their study. 
We included government press conferences as the third arena because they were important during the COVID-19 pandemic. They became a central venue for announcing new measures, with the health minister regularly appearing alongside leading scientific experts. framing the severity of the crisis and influencing subsequent media coverage \cite{hayek_2024}. 
While press conferences gained relevance  during the pandemic, we expect a moderate share of expert argumentation.

%But while several partially compatible and partially competing approaches of classifying arguments in political discourse exist, the prerequisite for applying these concepts to large text corpora is still not solved: So far, no convincing approach has been developed to  automatically identify and extract arguments in political texts, especially for cases where text spans containing arguments differ in length and may or may not comprise more than one sentence.

%\sebp{TODO for me: discuss approaches/tools for argumentative text analysis and argument mining. RST. RESEARCH GAP also discuss that much computational work on argumentation assumes relatively short and *primarily* argumentative texts while in political discourse, argumentative passages alternate with non-argumentative passages; no existing tools are really applicable to the types of texts we consider.}

\section{Dataset}
\subsection{Corpus Selection}
\label{sec:corpus-selection}
Our dataset of political discourse in the German political sphere consists of plenary speeches (Bundestagsreden), health committee protocols (Gesundheitsausschussprotokolle), and government press conferences (Bundespressekonferenzen). In this way, we aim to capture the full spectrum of parliamentary discursive arenas for one topical domain. Plenary speeches represent politicized front-stage deliberation and political competition. The health committee is a parliamentary back stage. And government press conferences are an interface between the general public and the parliament.

While plenary speeches are held on the agenda set in parliament and are delivered in monologue form, health committee and press conferences are dialogical formats, mostly organised as Q\&A sessions. Press conferences usually begin with statements by the participants, after which the government spokesperson responds to questions. The health committee is more or less scripted, as each party can invite (scientific) experts and question them according to the set agenda.

Our sampling strategy was determined in advance, as we sought to capture all COVID-related parliamentary discourse. Or period of observation starts at the beginning of 2020 with the onset of the pandemic and ends in April 2023 after all pandemic measures had ended and Health Minister Karl Lauterbach officially declared the pandemic over. We used the GermaParl Corpus \cite{Blaette_Leonhardt_2023} as data source for plenary speeches.  The protocols of all public meetings of the health committee were downloaded from the archived websites of each legislative period (\url{https://www.bundestag.de/ausschuesse/gesundheit}) and  the written transcripts of the Bundespressekonferenzen were obtained from \url{jungundnaiv.de}, a German journalist platform that provides complete coverage of these press conferences \cite{Jung_und_Naiv_BPK}.

\subsection{Preprocessing}
The resulting dataset contains 118,745 speeches, 205 protocols and 512 press conferences. We then filtered the dataset for COVID-19-related content using a dictionary of 1,958 keywords, which was using terms from the Leibniz Institute for the German Language relating to the COVID-19 pandemic \cite{Leibniz_Institute_COVID_Wordlist}. We refined our approach with regular expression patterns to capture all relevant text segments and removed all terms not related to the medical domain. These are mostly terms from socially triggered discourse without medical import, such as ‘click and collect’, ‘Drosten-Ultra’ or ‘Wiesn-Party’. This resulted in a total of 9,737 COVID-related texts (9,299 speeches, 159 protocols, 279 press conferences). 
%We assigned a unique text ID to each annotated text. 
We added metadata such as the speaker’s first and last name, date, party, and session.

\subsection{Annotation Schema}

Following \citet[p.~8]{Nordin_Schiappa_2024} we define our core concept, an \textit{argument}, as a text segment consisting of one or more sentences and comprising a claim and its justification. Claims are statements that suggest an action, such as a demand, plan, position, or recommendation. Our definition thus represents a variant of the definition of arguments as claims supported by reasons.

%\sebp{Here people might ask why we don't follow any of the existing argument(ation) annotation schemas from the argument mining literature. Can we make it clearer what theoretical framework this belongs to?}
Rather than adopting annotation schemes from the argument mining literature, our approach is grounded in the framework of practical argumentation and informal logic, particularly drawing on the Toulmin model. Initial attempts to operationalize speech act classification proved insufficient, as our first annotation sample did not align with the speech act categories. We therefore adopted a claim-based understanding of argumentation, showing how political claims are justified, rather than identifying argumentative units. While argument mining primarily focuses on the identification and the reconstruction of arguments, our aim is to capture modes of practical argumentation in political discourse. Specifically, we aim at classifying different modes of justification for political claims. 

To further distinguish these different types of justifications, we used a manual annotation of a subset of texts as an inductive baseline to assess whether stable categories could be derived. 
%Based on this, we developed a classification scheme. 
Building on this foundation, we developed a classification scheme aimed at capturing systematic differences in argumentative styles. We sought to differentiate between expert-driven and layperson's argumentation. Therefore we defined three basic types: \textit{evidence-based (disciplinary) justifications}, which rely on data, studies, models, or measurable evidence; \textit{normative and pragmatic justifications}, which are based on values, norms, feasibility, or utility; and \textit{ideological and analogical justifications}, which are based on opinions, rhetoric, or anecdotal and selective reasoning without verifiable evidence \cite{Toulmin_2003, Reyes_2011, van_Dijk_2000}. To capture expert argumentation, we assumed that only evidence-based (disciplinary) arguments qualify as expert arguments and added a binary indicator for whether the justification falls within the  subject area. The annotation scheme was further extended to distinguish between medical and non-medical evidence-based justifications. 

We sought to distinguish the COVID-19 discourse thematically in order to examine whether specific subtopics of the debate offer increased expert argumentation. We therefore classified each argument according to its subject. We used the manual annotation as a baseline and further refined it into six main categories: \textit{healthcare} (treatment, access, personnel), \textit{containment measures} (e.g. contact reduction, testing), \textit{social issues} (socio-economic issues, education, rights), \textit{research} (methods, data, expertise), international issues (cross-border issues), and \textit{logistics} (implementation, administration, communication, funding).

\begin{table*}[t]
\centering
\small
\begin{tabular}{p{0.38\linewidth} p{0.25\linewidth}p{0.30\linewidth}}
\hline
\textbf{Evidence-based Argument} & \textbf{Normative Argument}  & \textbf{Ideological Argument}\\
\hline
\textit{“Strict contact restrictions are necessary because they can halt and reverse the spread of infection.”} &
\textit{“Something needs to be done. If we ignore the problem now, the situation will get worse.”}  &\textit{“It seems to be noone from Ms. Merkel, Mr. Spahn and Mr. Altmaier knows - because they are childless - what they are demanding from our youth. It is a catastrophe.”}\\

\textbf{Claim:} “Strict contact restrictions are necessary [...]” &
\textbf{Claim:} “Something needs to be done.”  &
\textbf{Claim:}  “It is a catastrophe.“\\

\textbf{Topic:} Containment measures &
\textbf{Topic:} Unspecified  &\textbf{Topic:} Social\\

\textbf{Justification:} “[...] because they can halt and reverse the spread of infection.” &
\textbf{Justification:} “If we ignore the problem now, the situation will get worse.”  &\textbf{Justification:} "[...]because they are childless[...]"\\

\textbf{Type:} Evidence-based &
\textbf{Type:} Normative/pragmatic  &\textbf{Type:} Ideological/analogical\\

\textbf{Medical:} Yes &
\textbf{Medical:} No  &\textbf{Medical:} No\\
\hline
\end{tabular}
\caption{Agumentative spans and annotated labels.}
\label{tab:annotation_examples}
\end{table*}  

Table~\ref{tab:annotation_examples} provides English translations of two argumentative passages found in the corpus together with their annotation. The examples show how the annotation scheme distinguishes between  similarities and  differences in argumentative reasoning.  In both passages, a clear claim is supported by a justification, but the justifications differ. The first example represents an evidence-based argument, where the justification refers to an empirically testable mechanism. The second example, by contrast, exemplifies a normative argument, relying on a general appeal to urgency. 
%\joha{backup arg for table? “The strategy as well as quarantine and testing recommendations must be continuously adapted based on updated infection dynamics and viral load, since the scientific data on SARS-CoV-2 is constantly expanding.“
%(1) Claim: “The strategy as well as quarantine and testing recommendations must be continuously adapted[...]” 
%(2) Topic: Containment measures 
%(3) Justification: “[...] since the scientific data on SARS-CoV-2 is constantly expanding.”
%(3) Type: evidence-based
%(4) Medical: yes }

\subsection{Annotation Procedure}
\label{sec:annotation}
The annotation guidelines were presented to two annotators, both political science students, and first tested on a small sample. This formed part of the inductive annotation procedure, in which new documents were selected and uploaded each week. 
%The selection of texts followed a  sampling procedure to ensure a representation of all parties and to cover different phases of the COVID-19 pandemic over time. In addition, we pre-screened the material using the MARDY claim classifier in order to identify texts with a higher probability of containing argumentative statements.
To ensure a sufficient number of positive examples in our data, we pre-screen data for annotation using multiple factors. We sampled 
by date and political party, to cover different phases of the COVID-19 pandemic over time. We validated the selection by quantifying the presence of claims with an automatic classifier for detecting claims within German newspaper articles and manifestos \cite{blokker-etal-2020-swimming}. This classifier provides a lower bound for the presence of claims, due to the shift in text type which tends to decrease recall. Our analysis showed that approximately 65\% of the documents picked contained high number of claims, which we consider sufficient for our purposes.

%This analysis indicated that the our sample contains a substantial number of claims, and are therefore suitable to study argumentation. 
 % (higher % in BPK, lower % in GA)

The comparatively larger number of speeches is primarily due to structural differences. Speeches are substantially shorter and contain contributions by only a single speaker, whereas press conferences and committee protocols include multiple speakers and interaction sequences. The annotation effort per speech was approximately one quarter of the time required for the other documents.
 
The annotators coded the documents independently, while each document was annotated by both annotators, providing a reliable way to reduce subjectivity in the annotation process. Difficult cases were discussed after each annotation round to further refine the guidelines and assess their robustness. We used INCEpTION \citep{Klie_Bugert_Boullosa_EckartdeCastilho_Gurevych_2018} as annotation tool and configured the platform in accordance with the annotation guidelines by defining the annotation layers (ARG/JUST), features (topic for ARG; justification type and domain for JUST), and corresponding tagsets (topics for ARG; types and domains for JUST). As a result, the guidelines were accessible during annotation, and also structurally reflected in the annotation environment. The estimated annotation speed was roughly 23–30 arguments per hour, when we consider 15-25 min per speech and one hour for the other documents due to their length. The full (German) guidelines are available in Appendix~\ref{sec:appendix-guidelines}.

After the first annotation round, we analyzed the inter-coder agreement (ICA) with the Gamma statistic \citep{10.1162/COLI_a_00227}, a chance-corrected measure of agreement between sequence annotations that incorporates both labeling agreement (like Cohen's $\kappa$) and segmentation choice (which $\kappa$ does not take into account). Like for $\kappa$, its range is $[-1,1]$, where 1 indicates perfect agreement and 0 agreement at chance level.

As the second-to-last row of Table~\ref{tab:anno-stats} shows, ICA after initial annotation was mediocre, mainly due to differences in the identification of argument boundaries. We therefore created a gold standard in iterative steps, similar to other semantic annotation projects (e.g., OntoNotes, \citealt{hovy-etal-2006-ontonotes}). First, matching annotations were merged in the INCEpTION tool. Second, arguments and justifications without matching codes were manually reviewed by a third annotator with advanced expertise in political science and refined in accordance with the codebook. Thereafter, the ICA between the created gold standard and the initial annotations was calculated (final row of Table~\ref{tab:anno-stats}). This resulted in $\gamma$ scores around 0.7, reflecting the success of the adjudication procedure and demonstrating that the gold standard indeed represents a consensus between the two annotations. Still, we take away that it is hard to get annotators to agree perfectly on argument spans.

We do not carry out more detailed analysis of category correspondences, since these are hard to align in a combined segmentation-and-categorization task.

%Most disagreements concerned the annotation of the subject of an argument.\sebp{Can this maybe shown in one of the examples above?} Differences in argument length were the second most common issue, followed by arguments identified by only one annotator. These single annotations were reviewed individually and were often confirmed. 
%\nina{Table 2, Speeches: arg spans/(sentences/100) are not =arg/100. deviation feels quite large for a rounding error? Also there should be 15 protocols?}
%\sebp{@Nina, numbers correct?}
%\sebp{And then how was the final gold standard created?}
%\sebp{So the long and short of it is that it doesn't really make sense to compute/report inter-coder agreement here, right?} \joha{Yes and No, we might need the Gamma Data from Nina, so we can say whether they ICA was okay in the first place. if we use the Krippendorf Data from INCEpTION, we rather not tell this story here, since it does not report our main interests well enough, due to its calculation on perfectly correct span overlaps}\nina{ICA for BT: 0.46, GA: 0.37, BPK: 0.39, increased to ~0.64-0.8 when compared with curation (arguments&justifications in full texts considered, but similar results with only arguments. less improvement on paragraph-level)}

\begin{table}[tb!]
\centering
\resizebox{\columnwidth}{!}{%
\begin{tabular}{lrrrr}
\toprule
 & BT & GA & BPK & Total \\
\cmidrule(lr){2-5}
Documents & 60 & 15 & 12 & 87 \\
Contributions & 60 & 1046 & 1450 & 2465 \\
Sentences & 1881 & 9350 & 5888 & 17119 \\
Avg. Sent./Text & 31.35 & 623.33 & 490.67 & 196.77 \\
ARG Spans & 409 & 1079 & 201 & 1689 \\
ARG / 100 Sent. & 21.74 & 11.54 & 3.41 & 9.87 \\
JUST Spans & 457 & 1245 & 216 & 1918 \\
JUST / 100 Sent. & 24.30 & 13.32 & 3.67 & 11.20 \\
\cmidrule(lr){2-5}
Initial ICA ($\gamma$) & 0.46 & 0.37 & 0.39 \\
Avg. ICA with Gold ($\gamma$) & 0.72 & 0.69 & 0.73 \\
\bottomrule
\end{tabular}%
}
\caption{Descriptive statistics by arena: BT (Bundestag speeches), GA (Gesundheitsausschuss protocols), BPK (Bundespressekonferenz press conference)}
\label{tab:anno-stats} 
\end{table}

\subsection{Corpus Findings}

Our corpus consists of 9,299 parliamentary speeches, 159 committee protocols, and 279 press conferences, for a total of 9,737 texts. Note that the two latter types collapse multiple contributions in one document and are therefore much longer. 
Of these, we fully annotated 87 documents, with an average of 19.47  argument spans per documents.

%\sebp{Sounds like a very skewed distribution with the speeches taking the lion's share. Is that true, or are the speeches much shorter than the other documents?}\joha{Yes, speeches are much shorter} 
%\sebp{Ich habe 'Text' durch 'Document' und 'Statement' durch 'Contribution' ersetzt -- 'statement' wird in der NLP oft als synonym zu 'sentence' verwendet}
 %\sebp{Wir haben viel mehr BT ausgewählt (60/87) als von den anderen Arenen - vermutlich weil BT-Dokumente viel kürzer sind. War das ein formaler Auswahlprozess, oder eher Pi mal Daumen?}
%\joha{Auswahl erfolgt anhand Sprecher bzw. Parteien, Zeit Und Wahrscheinlichkeit für Argumente auf Grundlage des MARDY Claim Classifiers. Da die Dauer für die Annotation der Texte kürzer war, haben wir dort mehr Dokumente ausgewählt }

Table \ref{tab:anno-stats} shows statistics for the resulting annotated corpus, both in total and 
broken down by arena.
We identified 1,689 argument spans (ARG) and 1,918 justification spans (JUST), amounting to 3,607 spans overall. On average, this corresponds to 18.35 ARG spans per 100 sentences and 20.65 JUST spans per 100 sentences. The ARG spans were differentiated by topic, while the JUST spans were differentiated by justification type.

Looking at the topic distribution, logistics accounts for 60.1\%, followed by healthcare at 15.1\%. Social issues amount to 10.6\%, and containment measures to 8.3\%, while international issues account for 2.8\% and research for 2.8\%.

Regarding justification types, evidence-based justifications make up only 7.1\% of all JUST spans, while normative and pragmatic justifications account for 88.4\% and ideological/analogical justifications for 4.4\%.
Among the evidence-based JUST spans, 61.5\% were classified as medical and 38.5\% as non-medical.
The sources also differ in their argument density. Speeches show the highest average, with 23.02 ARG and JUST spans per text, followed by protocols (12.43). Press conferences contain only 3.54 ARG and JUST spans per text.

Overall, most JUST spans appear to be normatively driven justifications. However, when looking at the share of evidence-based justifications, BPK shows the highest proportion at 16.7\%. Speeches account for 6.6\% and protocols for 5.7\%.

The data reveal a discipline-specific concentration of evidence-based arguments, which is in line with our assumption that expert argumentation primarily takes place in evidence-based arguments. %\sebp{Worauf bezieht sich 'in this category'?}
The observed argument density also corresponds to our expectations: Although plenary speeches are a central site of argumentation, the Q\&A format of press conferences appears to constrain argument density, with committee protocols falling between the other two sources. The press conferences appear to be the primary venue for expert argumentation, while plenary speeches are driven by normative justification.  Contrary to expectations, the health committee is only third.

These findings align directly with our expectation that appeals to expertise are context dependent but contrast with the expectation that such forms would be concentrated in committees, where argument density is high and expert knowledge might be expected most strongly. Although previous research focuses on public “frontstage” debate versus expertise-oriented “backstage” committee deliberation, we observe that expert argument appears more often in government press conferences, rather than in committee deliberation \cite{Karlsson_Persson_Martensson_2024,Andone_2016}. Although committees show a dense argumentative interaction, expertise-based justification appears most prominently in press conferences. In plenary speeches, we observe more normative reasoning: Justifications rest on general appeals to collective protection rather than on empirical evidence. For example, one MP argues that restrictions on fundamental rights are necessary to enable containment measures aimed at protecting the population.\footnote{Example: "Ja, zum Schutz der Bürger vor Corona sind Grundrechtseingriffe erforderlich, etwa durch Hygiene\-konzepte oder Abstandsmaßnahmen."}%\sebp{Beispiel / Verweis auf Tab 1?} 

The prominence of expert arguments in government press conferences may reflect their institutional logic: Arguments in press conferences are used to communicate governmental decisions in a formalized setting, where speakers usually first deliver a relatively short statement and then respond to the journalists’ questions. This setting seems to trigger expert-like justifications to provide reasonable and deliberative answers. Press conferences thus tend to feature dense and evidence-based reasoning, where statements e.g. about increasing incidence rates, the spread of the B.1.1.7 variant, and increasing ICU occupancy are used to justify concrete government decisions, reflecting a stronger reliance on empirically grounded argumentation.\footnote{Example: "Es gibt das Auftreten neuer Virusvarianten, das die epidemiologische Lage verändert hat. Es gibt insbesondere die deutlich ansteckendere Virusvariante B.1.1.7. Es gibt stetig steigende Inzidenzen. Wir liegen heute bei einem Wert von 140,9. Wir haben eben auch eine wieder stetig steigende Zahl von Menschen, die auf den Intensivstationen behandelt werden müssen. Deswegen hat das Kabinett heute eine Ergänzung des Infektionsschutzgesetzes beschlossen, und zwar, wie ich gesagt habe, als Formulierungshilfe für die Fraktionen von CDU/CSU und SPD." } 
%\sebp{Hier wären ein oder zwei Beispiele super! Oder ist eines der Beispiele in Tab 1 aus der BPK? Dann wäre hier ein Verweis super.}

In the health committee, experts provide descriptive input, e.g., when discussing staff vaccination rates. This leads to a high number of normative justifications. We attribute this to the interaction between politicians and invited experts, where justifications are more normative than evidence-based.\footnote{Example: "Ich habe im Moment bei den Mitarbeitern eine Impfquote von 98 Prozent. Zu keinem Zeitpunkt hatten wir größere Probleme, außer den üblichen, den Dienstplan zu gestalten. In Moment habe ich, wenn ich die gesamte Gruppe anschaue, sechs Stellen zu besetzen. Das sind also auch hier keine Horrorszenarien, wie sie immer wieder kommuniziert werden."}

Thus, we can see overall that expertise is not simply tied to institutional proximity to expert knowledge, but also to communicative function and audience. At the same time, the more strongly normative plenary speech aligns with theories of deliberative and representative politics that emphasize public justification, value contestation, and political positioning in plenary arenas \cite{Steenbergen_Baechtiger_Spoerndli_Steiner_2003}.

%each corpus separately in CSV format, allowing independent use with minimal additional processing.

\subsection{Text Preparation}
\label{sec:splitting}

For modeling use, we split the datasets individually into three parts: training, development, and test. We decided to make these parts equal-sized (33\% each), to accommodate the various possible approaches to modeling (traditional training/evaluation; pre-training and fine-tuning; zero-shot/few-shot LLM use) while keeping the test set large enough for robust evaluation.\footnote{The dataset is available at Zenodo: \url{https://doi.org/10.5281/zenodo.21718040}}

\section{Pilot Modeling Study}
\label{sec:modeling}

We now present the results of a pilot study limited to the automatic recognition of argumentative spans for the different arenas in our annotated corpus. From our perspective, this task is the first step in a future more comprehensive modeling endeavour that will also consider other aspects of the annotation that we have creating, including claim and justification spans, claim domains and justification types. We restrict ourselves to the argument recognition subtask since (a) it is the logically first step when facing unanalyzed political text: all other steps presuppose recognized arguments; (b) a reliable model for just this step would already play an enabling role in the analysis of argumentation in political texts across arenas (cf. Section \ref{sec:intro}).

\begin{figure}[tb!]
\small
\includegraphics[width=1.1\linewidth]{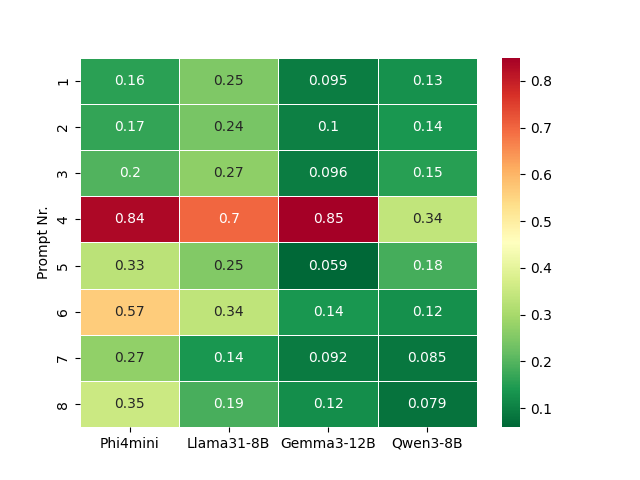}
\caption{Return rates of unfaithful quotations for combinations of models and prompt templates.}
\label{fig:prompt}
\end{figure}

\subsection{Model Choice} 

Since the models need to process German data, we require LLMs that have a good German proficiency. Monolingual German models are rare, and those that exist such as LLäMmlein \citep{pfister-etal-2025-llammlein} have undergone no or only basic instruction tuning. Given that our task is complex, we focus on multilingual instruction-tuned models. For reasons of practicality and reproducibility, we restricted ourselves to open-weights models in the range between 4B and 12B parameters \citep{liesenfeld_2023}. Specifically, we investigated Phi4-mini (4B, \citealt{microsoft2025phi4minitechnicalreportcompact}), Llama3.1 (8B, \citealt{grattafiori2024llama3herdmodels}),  Gemma3 (12B, \citealt{gemmateam2025gemma3technicalreport}) and Qwen3 (8B, \citealt{yang2025qwen3technicalreport}).

\subsection{Experimental design}
\label{sec:expt-design}
We follow the current standard usage mode for zero-shot text processing with large language models: we craft a prompt for our task, present the prompt with each input to LLMs, and parse the outputs.

\paragraph{Prompt selection.}
The model was instructed to return argumentative passages related to COVID from a given text. This is a task with a fairly complex output structure, namely a list of argumentative passages. Models can misquote the input, justify their output (contrary to instructions), or add arbitrary comments. 
Since we considered it unlikely that different LLMs would all work well with the same single prompt, we formulated a set of 8 prompt candidates, listed in Appendix \ref{sec:prompt-candidates}, that varied in length and detail of definition, closeness to guidelines, instructions on output form as well as language of instruction. These templates were evaluated on the train set. We then selected the prompt that led to the lowest number of unfaithful quotations from the input, across arenas. The results are shown in Figure~\ref{fig:prompt}. Note that this evaluation does not require gold standard annotation, only input text. The results show that one prompt (\#4) fails for all models and that models indeed prefer different prompts (e.g., Phi4mini prompt \#1 vs. prompt \#8 for Qwen3-8B). For the respective best prompt, the unfaithful quotation rate lies between 6\% (Gemma3) and 16\% (Phi4mini).

%The texts were presented to the LLMs by contributions (cf. Table~\ref{tab:anno-stats}).

\paragraph{Answer interpretation.} 
Our answer interpretation component accommodates variability in LLM output by trying to identify three types: (a), a list of arguments with quotation marks; (b), any other kind of list of arguments; (c) an empty list, typically accompanied with a justification. For (a), the text in-between quotation marks is extracted. For (b), we match a list with a regular expression. When some bullet point is recognized, everything located on the line is considered an argument. For (c), we look for keyphrases related to an empty return, such as \textit{"leere Liste"} or \textit{"keine argumentativen Textstellen"}. 
Outputs that remain unrecognized are treated as an empty list for the purposes of evaluation; however we keep track of the number of such 'invalid response' cases and report them separately.

\subsection{Evaluation Metrics}
We assess model predictions with a couple of evaluation metrics. First, we consider three variants of sequence-level F$_1$ score. All of these are computed at the level of annotation spans (not individual tokens) and differ in what they consider true positives (TPs): In the strict match condition, a model-predicted span is a TP only if it exactly matches a gold standard TP. In the inclusion condition, a model-predicted span is a TP if it is included in a gold standard span. In the partial match condition, each predicted span is aligned with the best-aligned gold span, and the TP count is the fraction of their overlap divided by predicted span length. Thus, strict match is a harsh criterion while partial match rewards models already for mostly wrong predictions. Inclusion attempts to strike a balance by rewarding models for not overpredicting spans.
%\sebp{@Nina, correct?} \nina{looks good!} %\nina{Yes. Also: for inclusion i took tp=1 if overlap is found, whereas for partial tp = token-ratio overlap. not sure if it is worth mentioning?}

The other metric we consider is Gamma ($\gamma$), which is usually used as a chance-corrected metric for inter-coder agreement (ICA) in sequence annotation (cf. Section \ref{sec:annotation}). Agreement between model-predicted spans and gold-standard spans with $\gamma$ can thus be compared to ICA numbers from above.

Finally, we report the percentage of invalid model responses, the number of non-argumentative contributions according to models and gold standard, and predicted argument density, measured as number of  arguments per 100 sentences.
%as in Table~\ref{tab:anno-stats}.

\subsection{Baselines}

As points of comparison for the LLM-based approach, we consider three simple baselines. Baseline 1 does not predict any arguments. Since this baseline has a recall of 0, its performance is always zero. Baseline 2 predicts that the whole text is a single argument. Since this baseline has a very low precision, its F1 score is also very low and we do not report detailed results. Baseline 3 predicts that every sentence is (its own) argument. This is the main baseline worth comparing against.

\input{figures/table_updated.tex}

%\begin{table*}[tb!h]
%\small
%\centering
%\begin{tabular}{llccc}
%\toprule
%& & BL 1: no arguments & BL2: text is one argument & BL 3: each sentence is one argument \\
%\midrule
%& & no arguments & whole contribution & each sentence \\
%& &  are found & is one argument & is an argument\\
%\midrule
%\multirow{3}{*}{BT} & Pr/Re/F1(strict) & 0.00 0.00 0.00  & 0.00 0.00 0.00 & 0.04 0.20 0.07 \\
% & Pr/Re/F1(inclusion) & 0.00 0.00 0.00 & 0.00 0.00 0.00 & 0.62 0.77 0.69\\
% & Pr/Re/F1(partial) & 0.00 0.00 0.00 & 0.19 0.03 0.05 & 0.67 0.79 0.73\\
%\midrule
%\multirow{3}{*}{GA} & Pr/Re/F1(strict) & 0.00 0.00 0.00 & 0.00 0.00 0.00 & 0.01 0.09 0.02\\
% & Pr/Re/F1(inclusion) & 0.00 0.00 0.00 & 0.00 0.00 0.00 & 0.51 0.80 0.62\\
% & Pr/Re/F1(partial) & 0.00 0.00 0.00 & 0.25 0.19 0.22 & 0.53 0.81 0.64\\
%\midrule
%\multirow{3}{*}{BPK} & Pr/Re/F1(strict) & 0.00 0.00 0.00 & 0.00 0.00 0.00 &0.00 0.07 0.01\\
% & Pr/Re/F1(inclusion) & 0.00 0.00 0.00 & 0.00 0.00 0.00 & 0.14 0.73 0.23\\% & Pr/Re/F1(partial) & 0.00 0.00 0.00 & 0.06 0.29 0.10 & 0.15 0.75 0.25\\
% \bottomrule
%\end{tabular}
%\caption{Performance of baselines on argumentative passage extraction in each arena.}
%\end{table*}

\subsection{Modeling Results}

The results for extracting argumentative passages from contributions in the three arenas are shown in Table \ref{tab:results}. We use the best prompt for each model (cf. Figure \ref{fig:prompt}). Our main observations are as follows.

\paragraph{Exact match F$_1$.} Since exact match is a harsh success criterion for the models,  results are in the single digits for almost all models. This is not surprising, given the difficulty of human annotators to agree on exact boundaries (cf. Section \ref{sec:annotation}) but underlines the difficulty to precisely delineate argumentative passages in our data. It is Gemma3, the largest model, which obtains the highest numbers on Bundestag and Gesundheitsausschuss.

\paragraph{Partial / inclusion F$_1$.} According to these more lenient metrics, the identification of argumentative passages is not impossible, but still a seriously challenging task. Phi4mini is the overall winner, showing robust performance across all three arenas. This is striking, given that it is the smallest among our LLMs (4B parameters). Gemma3 and Llama3.1 follow in second place for Bundestag and Gesundheitsausschuss, while QWen shows rather good performance on Bundespressekonferenz -- see below for an explanation.

\paragraph{Gamma scores.} The Gamma scores
indicate that some of the models perform close to or even at chance level, notably Llama3.1 and Gemma3 on Bundestag. This is  different for the two other arenas, where Gamma scores reach 0.35 (Gesundheitsausschuss) and 0.44 (Bundespressekonferenz).This is a level comparable to the agreement among human annotators (cf. Table~\ref{tab:anno-stats}).
 According this metric, QWen3 is the model with the overall most robust performance. 
 
%\sebp{@Nina any idea why?}\nina{Einige Ausgaben sind eigentich leere Liste - es ist sehr variabel, wie genau das Modell es ausdrückt. Da es ganz strickt nach keywords sucht, fällt es oftmals durch meine methode durch. Noch ein paar waren zwar Zitate, aber nicht an der Form als solche erkennbar z.B. bei Einzelargumenten. dort lässt das model wahrscheinlich die übliche 'bullet point form' weg, da es nur 1 ist. Kann gut sein, dass da bei BPK weniger Argumente vorkommen, sind solche Fälle öfter.}

\paragraph{Argument density.} The results above can be explained, to an extent, 
by the differences in argument density among arenas. According to Table~\ref{tab:anno-stats}, the gold standard argument densities range between 22 arguments per 100 sentences (Bundestag), 12 arguments per 100 sentences (Gesundheitsausschuss) and 3 arguments per 100 sentences (Bundespressekonferenz).
 We observe striking differences among models regarding the predicted argument density.  All models underpredict arguments on the Bundestag data (8--14 arg. / 100 sent.) but 
most mildly overpredict arguments on the Gesundheitausschuss and extremely overpredict on the Bundespressekonferenz (except QWen3). The most extreme case
is Gemma3 which predicts 26 arguments per 100 sentences on BPK, almost 10 times the correct density. This naturally results in low precision. In comparison, Phi4mini estimates the argument density most accurately on Bundestag and Gesundheitsausschuss, which accounts for its good performance on these corpora. Only on Bundespressekonferenz, its density estimate is off; here, it is QWen3 which comes closest.

The differences between assumed argument density across arenas is particularly striking since all of these models are zero-shot. The only contact that the models have had with the evaluation data is the prompt selection step (cf. Section~\ref{sec:expt-design}).

Across all models and arenas, the closer a model manages to get to the argument density in the gold standard, the better its predictions:  F1 scores (partial match) and density accuracy (difference between gold and predicted density, divided by gold density) are inversely correlated (Spearman's $\rho$=-0.56, $n$=12, $p$=0.06). 
% a <- read.csv("density.csv")
% a$dd <- abs(a$density-a$golddens) / a$golddens
% cor.test(a$f1, a$dd, method="spearman")
Arguably, the lower argument density of Bundes\-presse\-konferenz stems from the fact that about 80 percent (406/490) of the contributions are purely descriptive, with no argumentation present. As the numbers on non-argumentative contributions according to models and gold-standards show, this is a major stumbling block for some of the models.
%This appears to be an instance of the well-known confirmation bias in LLMs \citep{sicilia-etal-2025-accounting}, where models are unwilling to reject the presupposition of the prompt (in this case, that there are arguments in the text). 

%\paragraph{Invalid responses.} All models show acceptable levels of invalid responses (given our answer interpretation) despite the complex task, with a maximum of 5\% invalid responses. Qwen3-8B is the most compliant model, always yielding an interpretable answer, but model size does not appear to be the dominant factor. We find many more invalid responses for the Bundespressekonferenz arena, as a side-effect of the many argument-less contributions (compare previous paragraph).

%\nina{Qwen BPK: there was an error in one test point - the model got caught in an endless loop. current score excludes that point. On avg. args for EuroLLM/BPK: model actually did return  ~2-5 arguments/paragraph, while often citing the instruction}}

\paragraph{Baseline.} The baseline which assumes that every sentence is a separate argument predicts, by definition, an argument density of 100, and never returns an empty argument list. It performs surprisingly strongly. In terms of F1 scores, the baseline outperforms the best LLMs on the Bundestag and Gesundheitsausschuss arenas and does about as well as the median LLM on Bundespressekonferenz. However, the much smaller  Gamma scores on all arenas (even negative on BT and GA) provide evidence for the essentially random nature of the predictions. Again, the differences among
the arenas reflect their differences in argument density: a baseline which assumes that every sentence is an argument performs better for arenas with denser argumentation.

\section{Conclusions and Future Work}

This paper investigated the differences in argumentation patterns among different arenas in German political discourse during the COVID-19 pandemic. In the absence of existing corpora to  ground our analysis, we make two contributions: (a) manual annotation of a 17k-sentence corpus comprising three important political arenas with argumentative passages, justifications, and corresponding categories; and (b) a pilot study to automatically recognize argument boundaries as a first step towards automating the information that we annotated.

Our main insights are as follows. First, identifying the boundaries of argumentative passages in political discourse is difficult both for human annotators and for models. We also found lower agreement compared to previous studies \citep{haddadan-etal-2019-yes,poiaganova-stede-2025-debates}. The extent to which this is due to our formulation of the task as sequence identification (and not sentence-level classification) is a matter of future work. An alternative route would be to embrace the differences among annotators in the spirit of perspectivism \citep{falk-etal-2024-overview}.

Second, we found major differences in argumentation patterns among our three arenas: the Bundespressekonferenz, while having the lowest argument density, shows the highest proportion of domain expertise-related justifications, while in plenary speeches normative justifications dominate.% \sebp{Consequences?}

Third, a major contributing reason for the difficulty that zero-shot LLMs have in identifying argumentative passages is that they struggle with estimating the argument density in input texts. This is likely because the models are trained to accept the presuppositions of instructions (in our case, the presence of arguments in the documents) and thus are reluctant to dismiss these presuppositions. As a consequence, our LLMs found it hard to beat a simple 'everything is an argument' baseline on the Bundestag and Gesundheitsausschuss texts. In future work, we plan to explore the role of simple embedding-based classifiers to 'pre-screen' documents for argumentative content \citep{ruiz-dolz-lawrence-2023-detecting}.

Other modeling options that were out of scope for this first study is the use of few-shot setups which could provide LLMs with priors about argument density \citep{schaefer-etal-2022-selecting}; replacing LLMs with 'more traditional' embedding-based classifiers that have shown good performance for argument recognition in previous work \citep{poiaganova-stede-2025-debates}; and modeling the remaining aspects of our annotation (justification spans as well as argument and justification categories).

\section*{Limitations}

At the corpus level, our annotation effort is focused on a single domain (COVID-19). While we believe this to be a reasonable choice (cf. Section \ref{sec:corpus-selection}), we acknowledge that our findings may not generalize straightforwardly to other domains. Related to this, we only consider a single committee (the health committee -- Gesundheitsausschuss); investigations of other domains should consider other committees.

At the model level, we have considered five 'consumer-sizes' open-weights LLMs but considered neither simpler solutions (embedding-based models) nor very large properietary LLMs which might provide considerably better performance. Prompts were also selected from a small number of templates.

\bibliography{custom}

\appendix

\section{Full Annotation Guidelines}
\label{sec:appendix-guidelines}

(follow on next complete page)

\includepdf[pages=-]{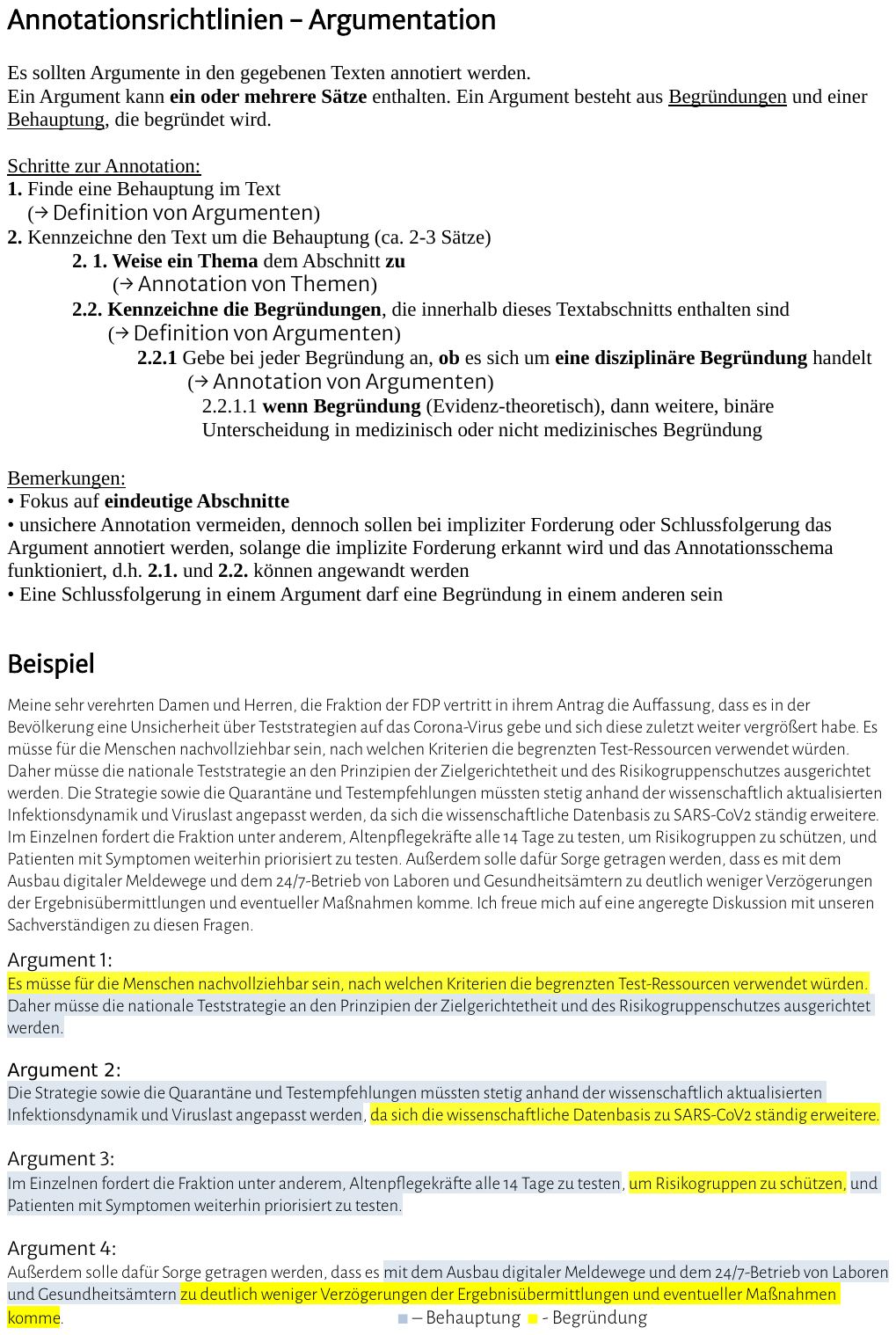}

\section{Prompt Candidates}
\label{sec:prompt-candidates}
%\nina{It would be better if the prompt numbers are changed to 1-8. I'll leave the old id-numbers until I changed them in the heatmap-figure.}
\begin{table}[ht!]
\small
\begin{tabular}{llp{12.5cm}} 
\toprule
System prompt & & You are a helpful AI assistant specialized on argument detection. You operate in German.\\ \midrule
Prompt & 1 & Finde argumentative Textstellen zu dem Thema Corona innerhalb des gegebenen Textabschnittes. Ein Argument besteht aus mindestens einem Satz und enthält eine Stellung und eine oder mehrere Begründungen. Es soll so kurz wie möglich gehalten werden und maximal 3 Sätze betragen. Ein Argument soll aus ganzen Sätzen bestehen. Es soll ein Bezug zu Corona haben. Gebe die Argumente als Liste aus. Gebe die Textstellen genauso aus, wie sie sind, ohne Veränderungen. Falls keine Argumente vorhanden sind, soll die Liste leer sein. Keine weiteren Kommentare oder Erkärungen. Finde die Argumente im folgendem Textabschnitt: \\ \cmidrule(lr){2-3}
& 2 & Finde argumentative Textstellen zu dem Thema Corona innerhalb des gegebenen Textabschnittes. Ein Argument besteht aus mindestens einem Satz und enthält eine Stellungsnahme und eine oder mehrere Begründungen. Es soll so kurz wie möglich gehalten werden und maximal 3 Sätze betragen. Ein Argument soll aus ganzen Sätzen bestehen. Es soll ein Bezug zu Corona haben. Gebe die Argumente als Liste aus. Gebe die Textstellen genauso aus, wie sie sind, ohne Veränderungen. Falls keine Argumente vorhanden sind, soll die Liste leer sein. Keine weiteren Kommentare oder Erkärungen. Finde die Argumente im folgendem Textabschnitt: \\ \cmidrule(lr){2-3}
& 3 & Finde argumentative Textstellen zu dem Thema Corona innerhalb des gegebenen Textabschnittes. Ein Argument besteht aus mindestens einem Satz und enthält eine Stellung und eine oder mehrere Begründungen. Es soll so kurz wie möglich gehalten werden und maximal 3 Sätze betragen. Ein Argument soll aus ganzen Sätzen bestehen. Die Stelle muss einen Bezug zu Corona haben. Das kann die Bereiche von Gesundheitsversorgung, Eindämmerungsmaßnahmen, Soziales, Forschung, Internationales oder Logistik betreffen. Insgesamt soll der Umgang mit der Epidemie-Situation oder Handlungen, die aus Pandemie entstanden sind, begründet werden. Gebe die Argumente als Liste aus. Gebe die Textstellen genauso aus, wie sie sind, ohne Veränderungen. Falls keine Argumente vorhanden sind, soll die Liste leer sein. Keine weiteren Kommentare oder Erkärungen. Finde die Argumente im folgendem Textabschnitt:\\ \cmidrule(lr){2-3}
& 4 & Es sollten argumentativen Stellen in dem gegebenen Text gefunden werden. Ein Argument kann ein oder mehrere Sätze enthalten. Ein Argument besteht aus Begründungen und einer Behauptung, die begründet wird. Eine Behauptung ist eine Aussage, die eine Handlung suggeriert. Z.B. Aufforderung, Vorhaben, Standpunkt, Empfehlung/Einschätzung oder Schlussfolgerung. Eine Begründung können Fakten und Bewertungen sein, oder auch z. B. bestehende Probleme, Abwertung anderer Vorschläge, Absicht, Prognosen, Vorteile oder (unabsichtigte Folgen), Veranschaulischungen, Verweise oder offenkundige Gründe sein. Ein Argument soll so kurz wie möglich gehalten werden und maximal 3 Sätze betragen. Die Stelle muss einen Bezug zu Corona haben. Es soll der Umgang mit der Epidemie-Situation oder Handlungen, die aus Pandemie entstanden sind, begründet werden. Gebe die Textstellen genauso aus, wie sie sind, ohne Veränderungen. Gebe die gefundenen Argumente als eine Liste aus: "Argument 1", "Argument 2", ... Falls keine Argumente vorhanden sind, soll de Liste leer sein: [] Finde die Argumente im folgendem Textabschnitt:\\ \cmidrule(lr){2-3}
& 5 & Find argumentative passages within given text. Arguments include a claim and at least one justification. Full reasoning should be included in one argumentative passage. However, the argument should be as short as possible and be 3 sentences max. Do not change any text within the arguments and return found passages as direct quotes. Return all arguments in a list of shape: "Argument 1", "Argument 2", ... If no arguments are found, return an empty list: [ ] Don't provide any comments or explanations. Find argumentative passages in the following text:\\ \cmidrule(lr){2-3}
& 6 & Es sollten argumentativen Stellen, die eine Behauptung mit Begründungen enthalten, in dem gegebenen Text gefunden werden. Ein Argument soll so kurz wie möglich gehalten werden, maximal 3 Sätze. Jede Stelle muss einen Bezug zu Corona haben. Gebe die Textstellen genauso aus, wie sie sind, ohne Veränderungen. Keine sonstigen Kommentare. Falls keine Argumente vorhanden sind, soll de Liste leer sein. Finde die Argumente im folgendem Textabschnitt:\\ \cmidrule(lr){2-3}
& 7 & Find argumentative passages within given text. Arguments include a claim and at least one justification. Full reasoning should be included in one argumentative passage. However, the argument should be as short as possible and be 3 sentences max. Do not change any text within the arguments and return found passages as direct quotes. If no arguments are found, return an empty list. Don't provide any comments or explanations. Find argumentative passages in the following text:\\ \cmidrule(lr){2-3}
& 8 & Find argumentative passages within given text. Arguments include a claim, e.g. a call to action, and at least one justification. Justification can be facts or assessment, for example. Full reasoning should be included in one argumentative passage. However, the argument should be as short as possible and be 3 sentences max. The argumentative passages should be relevant to the Corona discourse and about dealing with the pandemic. Do not change any text within the arguments and return found passages as direct quotes. If no arguments are found, return an empty list. Don't provide any comments or explanations. Find argumentative passages in the following text:\\
\bottomrule
\end{tabular}
\end{table}

\end{document}

%% file: figures/table_updated.tex
% table containing results for models using different prompts (decided using quotation rate on test set)
% model output was pre-filtered to contain only proper citations

\begin{table*}[tb!h]
\small
\centering
\begin{tabular}{llc|cccc}
\toprule
& &  Baseline 3 & Phi4mini & Llama3.1-8B & Gemma3-12B & QWen3-8B\\
\midrule
\multirow{6}{*}{BT} & Pr/Re/F1(strict) & 0.04 0.20 0.07 & 0.09 0.06 0.07 & 0.05 0.02 0.03 & 0.15 0.07 \textbf{0.10} & 0.13 0.04 0.07\\
 & Pr/Re/F1(inclusion) & 0.62 0.77 \textbf{0.69} & 0.70 0.33 \textbf{0.44} & 0.60 0.22 0.32 & 0.67 0.26 0.38 & 0.70 0.20 0.31\\
 & Pr/Re/F1(partial) & 0.67 0.79 \textbf{0.73} & 0.78 0.35 \textbf{0.49} & 0.73 0.25 0.37 & 0.79 0.29 0.43 & 0.80 0.22 0.35\\ 
 & Gamma (ICA) & -0.12 & 0.09 & 0.01 & 0.02 & \textbf{0.14} \\ \cmidrule{2-7}
 & invalid responses & & 0/20 & 0/20 & 0/20 & 0/20\\
 & \#pred/gold empty list & 0/0 & 0/0 & 0/0 & 0/0 & 2/0\\
 & avg. args/100 sent. & 100 & 14.4 & 10 & 10.8 & 8.1\\
\midrule
\multirow{6}{*}{GA} & Pr/Re/F1(strict) & 0.01 0.09 0.02 & 0.05 0.07 0.06 & 0.06 0.07 0.06 & 0.06 0.10 \textbf{0.07} & 0.11 0.04 0.06\\
 & Pr/Re/F1(inclusion) & 0.51 0.80 \textbf{0.62} & 0.66 0.51 \textbf{0.57} & 0.78 0.47 0.56 & 0.57 0.52 0.54 & 0.62 0.19 0.29\\
 & Pr/Re/F1(partial) & 0.53 0.81 \textbf{0.64} & 0.69 0.52 \textbf{0.59} & 0.74 0.48 0.58 & 0.60 0.53 0.57 & 0.71 0.21 0.33\\ 
 & Gamma (ICA) & -0.09 & 0.33 & \textbf{0.35} & 0.22 & 0.33 \\\cmidrule{2-7}
 & invalid responses & & 14/350 & 0/350 & 0/350 & 0/350\\
 & \#pred/gold empty list & 0/176 & 171/176 & 175/176 & 67/176 & 235/176\\
 & avg. args/100 sent. & 100 & 17.6 & 14 & 21 & 4.5\\
\midrule
\multirow{6}{*}{BPK} & Pr/Re/F1(strict) & 0.00 0.07 0.01 & 0.03 0.09 0.04 & 0.07 0.06 \textbf{0.07} & 0.01 0.04 0.01 & 0.07 0.07 \textbf{0.07}\\
 & Pr/Re/F1(inclusion) & 0.14 0.73 0.23 & 0.19 0.39 \textbf{0.25} & 0.15 0.32 0.20 & 0.13 0.48 0.20 & 0.25 0.23 0.24\\
 & Pr/Re/F1(partial) & 0.15 0.75 0.25 & 0.20 0.41 \textbf{0.27} & 0.17 0.34 0.22 & 0.14 0.50 0.21 & 0.26 0.24 0.25\\ 
& Gamma (ICA) & 0.17 & 0.43 & 0.42 & 0.28 & \textbf{0.44} \\ \cmidrule{2-7}
& invalid responses & & 56/490 & 16/490 & 0/490 & 3/490\\
 & \#pred/gold empty list & 0/406 & 363/406 & 331/406 & 164/406 & 379/406 \\
 & avg. args/100 sent. & 100 & 11.6 & 10.7 & 25.8 & 4.2\\  \bottomrule
\end{tabular}
\caption{Performance of LLMs on argumentative passage extraction in three arenas (BT, Bundestag; GA, Gesundheitsausschuss; BPK, Bundespressekonferenz). Highest results for each metric in each arena boldfaced. Baseline 3: Every sentence is its own argument.}
\label{tab:results}
\end{table*}

%% file: custom.bib
@inproceedings{ruiz-dolz-lawrence-2023-detecting,
    title = "Detecting Argumentative Fallacies in the Wild: Problems and Limitations of Large Language Models",
    author = "Ruiz-Dolz, Ramon  and
      Lawrence, John",
    editor = "Alshomary, Milad  and
      Chen, Chung-Chi  and
      Muresan, Smaranda  and
      Park, Joonsuk  and
      Romberg, Julia",
    booktitle = "Proceedings of the 10th Workshop on Argument Mining",
    month = dec,
    year = "2023",
    address = "Singapore",
    publisher = "Association for Computational Linguistics",
    url = "https://aclanthology.org/2023.argmining-1.1/",
    doi = "10.18653/v1/2023.argmining-1.1",
    pages = "1--10"
}

@inproceedings{schaefer-etal-2022-selecting,
    title = "On Selecting Training Corpora for Cross-Domain Claim Detection",
    author = "Schaefer, Robin  and
      Knaebel, Ren{\'e}  and
      Stede, Manfred",
    editor = "Lapesa, Gabriella  and
      Schneider, Jodi  and
      Jo, Yohan  and
      Saha, Sougata",
    booktitle = "Proceedings of the 9th Workshop on Argument Mining",
    month = oct,
    year = "2022",
    address = "Online and in Gyeongju, Republic of Korea",
    publisher = "International Conference on Computational Linguistics",
    url = "https://aclanthology.org/2022.argmining-1.17/",
    pages = "181--186"
}

@article{10.1162/COLI_a_00227,
    author = {Mathet, Yann and Widlöcher, Antoine and Métivier, Jean-Philippe},
    title = {The Unified and Holistic Method Gamma ($\gamma$) for Inter-Annotator Agreement Measure and Alignment},
    journal = {Computational Linguistics},
    volume = {41},
    number = {3},
    pages = {437-479},
    year = {2015},
    month = {09},
     issn = {0891-2017},
    doi = {10.1162/COLI_a_00227},
    url = {https://doi.org/10.1162/COLI_a_00227},
    eprint = {https://direct.mit.edu/coli/article-pdf/41/3/437/1806629/coli_a_00227.pdf},
}

@inproceedings{poiaganova-stede-2025-debates,
    title = "From Debates to Diplomacy: Argument Mining Across Political Registers",
    author = "Poiaganova, Maria  and
      Stede, Manfred",
    editor = "Chistova, Elena  and
      Cimiano, Philipp  and
      Haddadan, Shohreh  and
      Lapesa, Gabriella  and
      Ruiz-Dolz, Ramon",
    booktitle = "Proceedings of the 12th Argument mining Workshop",
    month = jul,
    year = "2025",
    address = "Vienna, Austria",
    publisher = "Association for Computational Linguistics",
    url = "https://aclanthology.org/2025.argmining-1.20/",
    doi = "10.18653/v1/2025.argmining-1.20",
    pages = "205--216",
    ISBN = "979-8-89176-258-9"
}

@inproceedings{pfister-etal-2025-llammlein,
    title = {{LL}{\"a}{M}mlein: Transparent, Compact and Competitive {G}erman-Only Language Models from Scratch},
    author = "Pfister, Jan  and
      Wunderle, Julia  and
      Hotho, Andreas",
    editor = "Che, Wanxiang  and
      Nabende, Joyce  and
      Shutova, Ekaterina  and
      Pilehvar, Mohammad Taher",
    booktitle = "Proceedings of the 63rd Annual Meeting of the Association for Computational Linguistics (Volume 1: Long Papers)",
    month = jul,
    year = "2025",
    address = "Vienna, Austria",
    publisher = "Association for Computational Linguistics",
    url = "https://aclanthology.org/2025.acl-long.111/",
    doi = "10.18653/v1/2025.acl-long.111",
    pages = "2227--2246",
    ISBN = "979-8-89176-251-0",
}

@misc{gemmateam2025gemma3technicalreport,
      title={Gemma 3 Technical Report}, 
      author={Kishwarya Kamath and Johan Ferret and Shreya Pathak and Nino Vieillard and Ramona Merhej and Sarah Perrin and Tatiana Matejovicova and Alexandre Ramé and Morgane Rivière and Louis Rouillard and Thomas Mesnard and Geoffrey Cideron and Jean-bastien Grill and Sabela Ramos and Edouard Yvinec and Michelle Casbon and Etienne Pot and Ivo Penchev and Gaël Liu and Francesco Visin and Kathleen Kenealy and Lucas Beyer and Xiaohai Zhai and Anton Tsitsulin and Robert Busa-Fekete and Alex Feng and Noveen Sachdeva and Benjamin Coleman and Yi Gao and Basil Mustafa and Iain Barr and Emilio Parisotto and David Tian and Matan Eyal and Colin Cherry and Jan-Thorsten Peter and Danila Sinopalnikov and Surya Bhupatiraju and Rishabh Agarwal and Mehran Kazemi and Dan Malkin and Ravin Kumar and David Vilar and Idan Brusilovsky and Jiaming Luo and Andreas Steiner and Abe Friesen and Abhanshu Sharma and Abheesht Sharma and Adi Mayrav Gilady and Adrian Goedeckemeyer and Alaa Saade and Alex Feng and Alexander Kolesnikov and Alexei Bendebury and Alvin Abdagic and Amit Vadi and András György and André Susano Pinto and Anil Das and Ankur Bapna and Antoine Miech and Antoine Yang and Antonia Paterson and Ashish Shenoy and Ayan Chakrabarti and Bilal Piot and Bo Wu and Bobak Shahriari and Bryce Petrini and Charlie Chen and Charline Le Lan and Christopher A. Choquette-Choo and CJ Carey and Cormac Brick and Daniel Deutsch and Danielle Eisenbud and Dee Cattle and Derek Cheng and Dimitris Paparas and Divyashree Shivakumar Sreepathihalli and Doug Reid and Dustin Tran and Dustin Zelle and Eric Noland and Erwin Huizenga and Eugene Kharitonov and Frederick Liu and Gagik Amirkhanyan and Glenn Cameron and Hadi Hashemi and Hanna Klimczak-Plucińska and Harman Singh and Harsh Mehta and Harshal Tushar Lehri and Hussein Hazimeh and Ian Ballantyne and Idan Szpektor and Ivan Nardini and Jean Pouget-Abadie and Jetha Chan and Joe Stanton and John Wieting and Jonathan Lai and Jordi Orbay and Joseph Fernandez and Josh Newlan and Ju-yeong Ji and Jyotinder Singh and Kat Black and Kathy Yu and Kevin Hui and Kiran Vodrahalli and Klaus Greff and Linhai Qiu and Marcella Valentine and Marina Coelho and Marvin Ritter and Matt Hoffman and Matthew Watson and Mayank Chaturvedi and Michael Moynihan and Min Ma and Nabila Babar and Natasha Noy and Nathan Byrd and Nick Roy and Nikola Momchev and Nilay Chauhan and Noveen Sachdeva and Oskar Bunyan and Pankil Botarda and Paul Caron and Paul Kishan Rubenstein and Phil Culliton and Philipp Schmid and Pier Giuseppe Sessa and Pingmei Xu and Piotr Stanczyk and Pouya Tafti and Rakesh Shivanna and Renjie Wu and Renke Pan and Reza Rokni and Rob Willoughby and Rohith Vallu and Ryan Mullins and Sammy Jerome and Sara Smoot and Sertan Girgin and Shariq Iqbal and Shashir Reddy and Shruti Sheth and Siim Põder and Sijal Bhatnagar and Sindhu Raghuram Panyam and Sivan Eiger and Susan Zhang and Tianqi Liu and Trevor Yacovone and Tyler Liechty and Uday Kalra and Utku Evci and Vedant Misra and Vincent Roseberry and Vlad Feinberg and Vlad Kolesnikov and Woohyun Han and Woosuk Kwon and Xi Chen and Yinlam Chow and Yuvein Zhu and Zichuan Wei and Zoltan Egyed and Victor Cotruta and Minh Giang and Phoebe Kirk and Anand Rao and Kat Black and Nabila Babar and Jessica Lo and Erica Moreira and Luiz Gustavo Martins and Omar Sanseviero and Lucas Gonzalez and Zach Gleicher and Tris Warkentin and Vahab Mirrokni and Evan Senter and Eli Collins and Joelle Barral and Zoubin Ghahramani and Raia Hadsell and Yossi Matias and D. Sculley and Slav Petrov and Noah Fiedel and Noam Shazeer and Oriol Vinyals and Jeff Dean and Demis Hassabis and Koray Kavukcuoglu and Clement Farabet and Elena Buchatskaya and Jean-Baptiste Alayrac and Rohan Anil and Dmitry and Lepikhin and Sebastian Borgeaud and Olivier Bachem and Armand Joulin and Alek Andreev and Cassidy Hardin and Robert Dadashi and Léonard Hussenot},
      year={2025},
      eprint={2503.19786},
      archivePrefix={arXiv},
      primaryClass={cs.CL},
      url={https://arxiv.org/abs/2503.19786}, 
}

@misc{yang2025qwen3technicalreport,
      title={Qwen3 Technical Report}, 
      author={An Yang and Anfeng Li and Baosong Yang and Beichen Zhang and Binyuan Hui and Bo Zheng and Bowen Yu and Chang Gao and Chengen Huang and Chenxu Lv and Chujie Zheng and Dayiheng Liu and Fan Zhou and Fei Huang and Feng Hu and Hao Ge and Haoran Wei and Huan Lin and Jialong Tang and Jian Yang and Jianhong Tu and Jianwei Zhang and Jianxin Yang and Jiaxi Yang and Jing Zhou and Jingren Zhou and Junyang Lin and Kai Dang and Keqin Bao and Kexin Yang and Le Yu and Lianghao Deng and Mei Li and Mingfeng Xue and Mingze Li and Pei Zhang and Peng Wang and Qin Zhu and Rui Men and Ruize Gao and Shixuan Liu and Shuang Luo and Tianhao Li and Tianyi Tang and Wenbiao Yin and Xingzhang Ren and Xinyu Wang and Xinyu Zhang and Xuancheng Ren and Yang Fan and Yang Su and Yichang Zhang and Yinger Zhang and Yu Wan and Yuqiong Liu and Zekun Wang and Zeyu Cui and Zhenru Zhang and Zhipeng Zhou and Zihan Qiu},
      year={2025},
      eprint={2505.09388},
      archivePrefix={arXiv},
      primaryClass={cs.CL},
      url={https://arxiv.org/abs/2505.09388}, 
}

@misc{grattafiori2024llama3herdmodels,
      title={The {Llama} 3 Herd of Models}, 
      author={Aaron Grattafiori and Abhimanyu Dubey and Abhinav Jauhri and Abhinav Pandey and Abhishek Kadian and Ahmad Al-Dahle and Aiesha Letman and Akhil Mathur and Alan Schelten and Alex Vaughan and Amy Yang and Angela Fan and Anirudh Goyal and Anthony Hartshorn and Aobo Yang and Archi Mitra and Archie Sravankumar and Artem Korenev and Arthur Hinsvark and Arun Rao and Aston Zhang and Aurelien Rodriguez and Austen Gregerson and Ava Spataru and Baptiste Roziere and Bethany Biron and Binh Tang and Bobbie Chern and Charlotte Caucheteux and Chaya Nayak and Chloe Bi and Chris Marra and Chris McConnell and Christian Keller and Christophe Touret and Chunyang Wu and Corinne Wong and Cristian Canton Ferrer and Cyrus Nikolaidis and Damien Allonsius and Daniel Song and Danielle Pintz and Danny Livshits and Danny Wyatt and David Esiobu and Dhruv Choudhary and Dhruv Mahajan and Diego Garcia-Olano and Diego Perino and Dieuwke Hupkes and Egor Lakomkin and Ehab AlBadawy and Elina Lobanova and Emily Dinan and Eric Michael Smith and Filip Radenovic and Francisco Guzmán and Frank Zhang and Gabriel Synnaeve and Gabrielle Lee and Georgia Lewis Anderson and Govind Thattai and Graeme Nail and Gregoire Mialon and Guan Pang and Guillem Cucurell and Hailey Nguyen and Hannah Korevaar and Hu Xu and Hugo Touvron and Iliyan Zarov and Imanol Arrieta Ibarra and Isabel Kloumann and Ishan Misra and Ivan Evtimov and Jack Zhang and Jade Copet and Jaewon Lee and Jan Geffert and Jana Vranes and Jason Park and Jay Mahadeokar and Jeet Shah and Jelmer van der Linde and Jennifer Billock and Jenny Hong and Jenya Lee and Jeremy Fu and Jianfeng Chi and Jianyu Huang and Jiawen Liu and Jie Wang and Jiecao Yu and Joanna Bitton and Joe Spisak and Jongsoo Park and Joseph Rocca and Joshua Johnstun and Joshua Saxe and Junteng Jia and Kalyan Vasuden Alwala and Karthik Prasad and Kartikeya Upasani and Kate Plawiak and Ke Li and Kenneth Heafield and Kevin Stone and Khalid El-Arini and Krithika Iyer and Kshitiz Malik and Kuenley Chiu and Kunal Bhalla and Kushal Lakhotia and Lauren Rantala-Yeary and Laurens van der Maaten and Lawrence Chen and Liang Tan and Liz Jenkins and Louis Martin and Lovish Madaan and Lubo Malo and Lukas Blecher and Lukas Landzaat and Luke de Oliveira and Madeline Muzzi and Mahesh Pasupuleti and Mannat Singh and Manohar Paluri and Marcin Kardas and Maria Tsimpoukelli and Mathew Oldham and Mathieu Rita and Maya Pavlova and Melanie Kambadur and Mike Lewis and Min Si and Mitesh Kumar Singh and Mona Hassan and Naman Goyal and Narjes Torabi and Nikolay Bashlykov and Nikolay Bogoychev and Niladri Chatterji and Ning Zhang and Olivier Duchenne and Onur Çelebi and Patrick Alrassy and Pengchuan Zhang and Pengwei Li and Petar Vasic and Peter Weng and Prajjwal Bhargava and Pratik Dubal and Praveen Krishnan and Punit Singh Koura and Puxin Xu and Qing He and Qingxiao Dong and Ragavan Srinivasan and Raj Ganapathy and Ramon Calderer and Ricardo Silveira Cabral and Robert Stojnic and Roberta Raileanu and Rohan Maheswari and Rohit Girdhar and Rohit Patel and Romain Sauvestre and Ronnie Polidoro and Roshan Sumbaly and Ross Taylor and Ruan Silva and Rui Hou and Rui Wang and Saghar Hosseini and Sahana Chennabasappa and Sanjay Singh and Sean Bell and Seohyun Sonia Kim and Sergey Edunov and Shaoliang Nie and Sharan Narang and Sharath Raparthy and Sheng Shen and Shengye Wan and Shruti Bhosale and Shun Zhang and Simon Vandenhende and Soumya Batra and Spencer Whitman and Sten Sootla and Stephane Collot and Suchin Gururangan and Sydney Borodinsky and Tamar Herman and Tara Fowler and Tarek Sheasha and Thomas Georgiou and Thomas Scialom and Tobias Speckbacher and Todor Mihaylov and Tong Xiao and Ujjwal Karn and Vedanuj Goswami and Vibhor Gupta and Vignesh Ramanathan and Viktor Kerkez and Vincent Gonguet and Virginie Do and Vish Vogeti and Vítor Albiero and Vladan Petrovic and Weiwei Chu and Wenhan Xiong and Wenyin Fu and Whitney Meers and Xavier Martinet and Xiaodong Wang and Xiaofang Wang and Xiaoqing Ellen Tan and Xide Xia and Xinfeng Xie and Xuchao Jia and Xuewei Wang and Yaelle Goldschlag and Yashesh Gaur and Yasmine Babaei and Yi Wen and Yiwen Song and Yuchen Zhang and Yue Li and Yuning Mao and Zacharie Delpierre Coudert and Zheng Yan and Zhengxing Chen and Zoe Papakipos and Aaditya Singh and Aayushi Srivastava and Abha Jain and Adam Kelsey and Adam Shajnfeld and Adithya Gangidi and Adolfo Victoria and Ahuva Goldstand and Ajay Menon and Ajay Sharma and Alex Boesenberg and Alexei Baevski and Allie Feinstein and Amanda Kallet and Amit Sangani and Amos Teo and Anam Yunus and Andrei Lupu and Andres Alvarado and Andrew Caples and Andrew Gu and Andrew Ho and Andrew Poulton and Andrew Ryan and Ankit Ramchandani and Annie Dong and Annie Franco and Anuj Goyal and Aparajita Saraf and Arkabandhu Chowdhury and Ashley Gabriel and Ashwin Bharambe and Assaf Eisenman and Azadeh Yazdan and Beau James and Ben Maurer and Benjamin Leonhardi and Bernie Huang and Beth Loyd and Beto De Paola and Bhargavi Paranjape and Bing Liu and Bo Wu and Boyu Ni and Braden Hancock and Bram Wasti and Brandon Spence and Brani Stojkovic and Brian Gamido and Britt Montalvo and Carl Parker and Carly Burton and Catalina Mejia and Ce Liu and Changhan Wang and Changkyu Kim and Chao Zhou and Chester Hu and Ching-Hsiang Chu and Chris Cai and Chris Tindal and Christoph Feichtenhofer and Cynthia Gao and Damon Civin and Dana Beaty and Daniel Kreymer and Daniel Li and David Adkins and David Xu and Davide Testuggine and Delia David and Devi Parikh and Diana Liskovich and Didem Foss and Dingkang Wang and Duc Le and Dustin Holland and Edward Dowling and Eissa Jamil and Elaine Montgomery and Eleonora Presani and Emily Hahn and Emily Wood and Eric-Tuan Le and Erik Brinkman and Esteban Arcaute and Evan Dunbar and Evan Smothers and Fei Sun and Felix Kreuk and Feng Tian and Filippos Kokkinos and Firat Ozgenel and Francesco Caggioni and Frank Kanayet and Frank Seide and Gabriela Medina Florez and Gabriella Schwarz and Gada Badeer and Georgia Swee and Gil Halpern and Grant Herman and Grigory Sizov and Guangyi and Zhang and Guna Lakshminarayanan and Hakan Inan and Hamid Shojanazeri and Han Zou and Hannah Wang and Hanwen Zha and Haroun Habeeb and Harrison Rudolph and Helen Suk and Henry Aspegren and Hunter Goldman and Hongyuan Zhan and Ibrahim Damlaj and Igor Molybog and Igor Tufanov and Ilias Leontiadis and Irina-Elena Veliche and Itai Gat and Jake Weissman and James Geboski and James Kohli and Janice Lam and Japhet Asher and Jean-Baptiste Gaya and Jeff Marcus and Jeff Tang and Jennifer Chan and Jenny Zhen and Jeremy Reizenstein and Jeremy Teboul and Jessica Zhong and Jian Jin and Jingyi Yang and Joe Cummings and Jon Carvill and Jon Shepard and Jonathan McPhie and Jonathan Torres and Josh Ginsburg and Junjie Wang and Kai Wu and Kam Hou U and Karan Saxena and Kartikay Khandelwal and Katayoun Zand and Kathy Matosich and Kaushik Veeraraghavan and Kelly Michelena and Keqian Li and Kiran Jagadeesh and Kun Huang and Kunal Chawla and Kyle Huang and Lailin Chen and Lakshya Garg and Lavender A and Leandro Silva and Lee Bell and Lei Zhang and Liangpeng Guo and Licheng Yu and Liron Moshkovich and Luca Wehrstedt and Madian Khabsa and Manav Avalani and Manish Bhatt and Martynas Mankus and Matan Hasson and Matthew Lennie and Matthias Reso and Maxim Groshev and Maxim Naumov and Maya Lathi and Meghan Keneally and Miao Liu and Michael L. Seltzer and Michal Valko and Michelle Restrepo and Mihir Patel and Mik Vyatskov and Mikayel Samvelyan and Mike Clark and Mike Macey and Mike Wang and Miquel Jubert Hermoso and Mo Metanat and Mohammad Rastegari and Munish Bansal and Nandhini Santhanam and Natascha Parks and Natasha White and Navyata Bawa and Nayan Singhal and Nick Egebo and Nicolas Usunier and Nikhil Mehta and Nikolay Pavlovich Laptev and Ning Dong and Norman Cheng and Oleg Chernoguz and Olivia Hart and Omkar Salpekar and Ozlem Kalinli and Parkin Kent and Parth Parekh and Paul Saab and Pavan Balaji and Pedro Rittner and Philip Bontrager and Pierre Roux and Piotr Dollar and Polina Zvyagina and Prashant Ratanchandani and Pritish Yuvraj and Qian Liang and Rachad Alao and Rachel Rodriguez and Rafi Ayub and Raghotham Murthy and Raghu Nayani and Rahul Mitra and Rangaprabhu Parthasarathy and Raymond Li and Rebekkah Hogan and Robin Battey and Rocky Wang and Russ Howes and Ruty Rinott and Sachin Mehta and Sachin Siby and Sai Jayesh Bondu and Samyak Datta and Sara Chugh and Sara Hunt and Sargun Dhillon and Sasha Sidorov and Satadru Pan and Saurabh Mahajan and Saurabh Verma and Seiji Yamamoto and Sharadh Ramaswamy and Shaun Lindsay and Shaun Lindsay and Sheng Feng and Shenghao Lin and Shengxin Cindy Zha and Shishir Patil and Shiva Shankar and Shuqiang Zhang and Shuqiang Zhang and Sinong Wang and Sneha Agarwal and Soji Sajuyigbe and Soumith Chintala and Stephanie Max and Stephen Chen and Steve Kehoe and Steve Satterfield and Sudarshan Govindaprasad and Sumit Gupta and Summer Deng and Sungmin Cho and Sunny Virk and Suraj Subramanian and Sy Choudhury and Sydney Goldman and Tal Remez and Tamar Glaser and Tamara Best and Thilo Koehler and Thomas Robinson and Tianhe Li and Tianjun Zhang and Tim Matthews and Timothy Chou and Tzook Shaked and Varun Vontimitta and Victoria Ajayi and Victoria Montanez and Vijai Mohan and Vinay Satish Kumar and Vishal Mangla and Vlad Ionescu and Vlad Poenaru and Vlad Tiberiu Mihailescu and Vladimir Ivanov and Wei Li and Wenchen Wang and Wenwen Jiang and Wes Bouaziz and Will Constable and Xiaocheng Tang and Xiaojian Wu and Xiaolan Wang and Xilun Wu and Xinbo Gao and Yaniv Kleinman and Yanjun Chen and Ye Hu and Ye Jia and Ye Qi and Yenda Li and Yilin Zhang and Ying Zhang and Yossi Adi and Youngjin Nam and Yu and Wang and Yu Zhao and Yuchen Hao and Yundi Qian and Yunlu Li and Yuzi He and Zach Rait and Zachary DeVito and Zef Rosnbrick and Zhaoduo Wen and Zhenyu Yang and Zhiwei Zhao and Zhiyu Ma},
      year={2024},
      eprint={2407.21783},
      archivePrefix={arXiv},
      primaryClass={cs.AI},
      url={https://arxiv.org/abs/2407.21783}, 
}

@misc{microsoft2025phi4minitechnicalreportcompact,
      title={{Phi-4-Mini} Technical Report: Compact yet Powerful Multimodal Language Models via Mixture-of-{LoRA}s}, 
      author={Abdelrahman Abouelenin and Atabak Ashfaq and Adam Atkinson and Hany Awadalla and Nguyen Bach and Jianmin Bao and Alon Benhaim and Martin Cai and Vishrav Chaudhary and Congcong Chen and Dong Chen and Dongdong Chen and Junkun Chen and Weizhu Chen and Yen-Chun Chen and Yi-ling Chen and Qi Dai and Xiyang Dai and Ruchao Fan and Mei Gao and Min Gao and Amit Garg and Abhishek Goswami and Junheng Hao and Amr Hendy and Yuxuan Hu and Xin Jin and Mahmoud Khademi and Dongwoo Kim and Young Jin Kim and Gina Lee and Jinyu Li and Yunsheng Li and Chen Liang and Xihui Lin and Zeqi Lin and Mengchen Liu and Yang Liu and Gilsinia Lopez and Chong Luo and Piyush Madan and Vadim Mazalov and Arindam Mitra and Ali Mousavi and Anh Nguyen and Jing Pan and Daniel Perez-Becker and Jacob Platin and Thomas Portet and Kai Qiu and Bo Ren and Liliang Ren and Sambuddha Roy and Ning Shang and Yelong Shen and Saksham Singhal and Subhojit Som and Xia Song and Tetyana Sych and Praneetha Vaddamanu and Shuohang Wang and Yiming Wang and Zhenghao Wang and Haibin Wu and Haoran Xu and Weijian Xu and Yifan Yang and Ziyi Yang and Donghan Yu and Ishmam Zabir and Jianwen Zhang and Li Lyna Zhang and Yunan Zhang and Xiren Zhou},
      year={2025},
      eprint={2503.01743},
      archivePrefix={arXiv},
      primaryClass={cs.CL},
      url={https://arxiv.org/abs/2503.01743}, 
}

@inproceedings{daxenberger-etal-2017-essence,
    title = "What is the Essence of a Claim? {C}ross-Domain Claim Identification",
    author = "Daxenberger, Johannes  and
      Eger, Steffen  and
      Habernal, Ivan  and
      Stab, Christian  and
      Gurevych, Iryna",
    editor = "Palmer, Martha  and
      Hwa, Rebecca  and
      Riedel, Sebastian",
    booktitle = "Proceedings of the 2017 Conference on Empirical Methods in Natural Language Processing",
    month = sep,
    year = "2017",
    address = "Copenhagen, Denmark",
    publisher = "Association for Computational Linguistics",
    url = "https://aclanthology.org/D17-1218/",
    doi = "10.18653/v1/D17-1218",
    pages = "2055--2066"
}

@inproceedings{cabrio2018five,
  author    = {Elena Cabrio and Serena Villata},
  title     = {Five Years of Argument Mining: A Data-Driven Analysis},
  booktitle = {Proceedings of the Twenty-Seventh International Joint Conference on Artificial Intelligence (IJCAI 2018)},
  pages     = {5427--5433},
  year      = {2018},
  doi       = {10.24963/ijcai.2018/764}
}

@inproceedings{liesenfeld_2023,
  title       = {Opening up {ChatGPT}: Tracking openness, transparency, and accountability in instruction-tuned text generators},
  author      = {Liesenfeld, Andreas and Lopez, Alianda and Dingemanse, Mark},
  year        = 2023,
  booktitle   = {Proceedings of the 5th International Conference on Conversational User Interfaces},
  publisher   = {Association for Computing Machinery},
  pages       = {1–6},
  doi         = {10.1145/3571884.3604316},
}

@inbook{Baechtiger_Dryzek_Mansbridge_Warren_2018,address={Oxford},
 series={Oxford Handbooks},
 title={Deliberative Democracy. An Introduction},
 ISBN={978-0-19-106456-2},
 booktitle={The Oxford Handbook of Deliberative Democracy},
 publisher={Oxford University Press},
 author={Bächtiger, André and Dryzek, John S. and Mansbridge, Jane and Warren, Mark E.},
 year={2018},
 pages={1–31},
 collection={Oxford Handbooks} }

@article{Steffensmeier_Schenck-Hamlin_2008,
 title={Argument Quality in Public Deliberations},
 volume={45},
 ISSN={1051-1431},
 DOI={10.1080/00028533.2008.11821693},
 number={1},
 journal={Argumentation and Advocacy},
 publisher={Routledge},
 author={Steffensmeier, Timothy and Schenck-Hamlin, William},
 year={2008},
 month={June},
 pages={21–36} }

@article{Steenbergen_Baechtiger_Spoerndli_Steiner_2003,
 title={Measuring Political Deliberation: A Discourse Quality Index},
 volume={1},
 DOI={10.1057/palgrave.cep.6110002},
 number={1},
 journal={Comparative European Politics},
 author={Steenbergen, Marco R. and Bächtiger, André and Spörndli, Markus and Steiner, Jürg},
 year={2003},
 pages={21–48} }

@article{Reyes_2011,
 title={Strategies of legitimization in political discourse: From words to actions},
 volume={22},
 ISSN={0957-9265},
 DOI={10.1177/0957926511419927},
 number={6},
 journal={Discourse \& Society},
 publisher={SAGE Publications Ltd},
 author={Reyes, Antonio},
 year={2011},
 month={Nov},
 pages={781–807} }

@article{Karlsson_Persson_Martensson_2024,
 title={Do Members of Parliament Express More Opposition in the Plenary than in the Committee? {C}omparing Frontstage and Backstage Behaviour in Five National Parliaments},
 volume={77},
 ISSN={0031-2290},
 DOI={10.1093/pa/gsac016},
 number={1},
 journal={Parliamentary Affairs},
 author={Karlsson, Christer and Persson, Thomas and Mårtensson, Moa},
 year={2024},
 month={Jan},
 pages={173–195} }

@inproceedings{Wachsmuth_Naderi_Hou_Bilu_Prabhakaran_Thijm_Hirst_Stein_2017,
 address={Valencia, Spain},
 title={Computational Argumentation Quality Assessment in Natural Language},
 url={https://aclanthology.org/E17-1017/},
 booktitle={Proceedings of the 15th Conference of the European Chapter of the Association for Computational Linguistics: Volume 1, Long Papers},
 publisher={Association for Computational Linguistics},
 author={Wachsmuth, Henning and Naderi, Nona and Hou, Yufang and Bilu, Yonatan and Prabhakaran, Vinodkumar and Thijm, Tim Alberdingk and Hirst, Graeme and Stein, Benno},
 editor={Lapata, Mirella and Blunsom, Phil and Koller, Alexander},
 year={2017},
 month={Apr},
 pages={176–187} }

@inproceedings{Reinig_Rehbein_Ponzetto_2024,
 address={Torino, Italia},
 title={How to Do Politics with Words: Investigating Speech Acts in Parliamentary Debates},
 url={https://aclanthology.org/2024.lrec-main.727/},
 booktitle={Proceedings of the 2024 Joint International Conference on Computational Linguistics, Language Resources and Evaluation (LREC-COLING 2024)},
 publisher={ELRA and ICCL},
 author={Reinig, Ines and Rehbein, Ines and Ponzetto, Simone Paolo},
 editor={Calzolari, Nicoletta and Kan, Min-Yen and Hoste, Veronique and Lenci, Alessandro and Sakti, Sakriani and Xue, Nianwen},
 year={2024},
 month={May},
 pages={8287–8300} }

@article{Andone_2016,
 title={Argumentative Patterns in the Political Domain: The Case of {E}uropean Parliamentary Committees of Inquiry},
 volume={30},
 ISSN={1572-8374},
 DOI={10.1007/s10503-015-9372-4},
 number={1},
 journal={Argumentation},
 author={Andone, Corina},
 year={2016},
 month={Mar},
 pages={45–60} }

@inproceedings{falk-etal-2024-overview,
    title = "Overview of {P}erspective{A}rg2024 The First Shared Task on Perspective Argument Retrieval",
    author = "Falk, Neele  and
      Waldis, Andreas  and
      Gurevych, Iryna",
    editor = "Ajjour, Yamen  and
      Bar-Haim, Roy  and
      El Baff, Roxanne  and
      Liu, Zhexiong  and
      Skitalinskaya, Gabriella",
    booktitle = "Proceedings of the 11th Workshop on Argument Mining (ArgMining 2024)",
    month = aug,
    year = "2024",
    address = "Bangkok, Thailand",
    publisher = "Association for Computational Linguistics",
    url = "https://aclanthology.org/2024.argmining-1.14/",
    doi = "10.18653/v1/2024.argmining-1.14",
    pages = "130--149"
}

@techreport{Blaette_Leonhardt_2023,
  author={Blaette, Andreas and Leonhardt, Christoph},
  title={{GermaParl} Corpus of Plenary Protocols},
  institution  = {zenodo},
  version={v2.0.1},
  year={2023},
  publisher={Zenodo},
  doi={10.5281/zenodo.10416536},
  url={https://doi.org/10.5281/zenodo.10416536},
  note={[Data set]}
}

@inproceedings{haddadan-etal-2019-yes,
    title = "Yes, we can! Mining Arguments in 50 Years of {US} Presidential Campaign Debates",
    author = "Haddadan, Shohreh  and
      Cabrio, Elena  and
      Villata, Serena",
    editor = "Korhonen, Anna  and
      Traum, David  and
      M{\`a}rquez, Llu{\'i}s",
    booktitle = "Proceedings of the 57th Annual Meeting of the Association for Computational Linguistics",
    month = jul,
    year = "2019",
    address = "Florence, Italy",
    publisher = "Association for Computational Linguistics",
    url = "https://aclanthology.org/P19-1463/",
    doi = "10.18653/v1/P19-1463",
    pages = "4684--4690"
}

@inproceedings{hovy-etal-2006-ontonotes,
    title = "{O}nto{N}otes: The 90{\%} Solution",
    author = "Hovy, Eduard  and
      Marcus, Mitchell  and
      Palmer, Martha  and
      Ramshaw, Lance  and
      Weischedel, Ralph",
    editor = "Moore, Robert C.  and
      Bilmes, Jeff  and
      Chu-Carroll, Jennifer  and
      Sanderson, Mark",
    booktitle = "Proceedings of the Human Language Technology Conference of the {NAACL}, Companion Volume: Short Papers",
    month = jun,
    year = "2006",
    address = "New York City, USA",
    publisher = "Association for Computational Linguistics",
    url = "https://aclanthology.org/N06-2015/",
    pages = "57--60"
}

@article{Lippi_Torroni_2016, title={Argument Mining from Speech: Detecting Claims in Political Debates}, volume={30}, url={https://ojs.aaai.org/index.php/AAAI/article/view/10384}, DOI={10.1609/aaai.v30i1.10384}, abstractNote={ &amp;lt;p&amp;gt; The automatic extraction of arguments from text, also known as argument mining, has recently become a hot topic in artificial intelligence. Current research has only focused on linguistic analysis. However, in many domains where communication may be also vocal or visual, paralinguistic features too may contribute to the transmission of the message that arguments intend to convey. For example, in political debates a crucial role is played by speech. The research question we address in this work is whether in such domains one can improve claim detection for argument mining, by employing features from text and speech in combination. To explore this hypothesis, we develop a machine learning classifier and train it on an original dataset based on the 2015 UK political elections debate. &amp;lt;/p&amp;gt; }, number={1}, journal={Proceedings of the AAAI Conference on Artificial Intelligence}, author={Lippi, Marco and Torroni, Paolo}, year={2016}, month={Mar.} }

@article{Visser2020Argumentation,
  author  = {J. Visser and B. Konat and R. Duthie and M. Koszowy and K. Budzynska and C. Reed },
  title   = {Argumentation in the 2016 US presidential elections: annotated corpora of television debates and social media reaction},
  journal = {Language Resources and Evaluation},
  volume  = {54},
  pages   = {123--154},
  year    = {2020},
  doi     = {10.1007/s10579-019-09446-8},
  url     = {https://doi.org/10.1007/s10579-019-09446-8}
}

@misc{Leibniz_Institute_COVID_Wordlist,
  author={{Leibniz IDS}},
  title={Word list for the {COVID-19} pandemic},
  url={https://www.owid.de/docs/neo/listen/corona.jsp},
  year = {undated},
  note={Accessed 25.06.2025}
}

@misc{Jung_und_Naiv_BPK,
  author={{Jung \& Naiv}},
  title={Bundespressekonferenz transcripts},
  url={https://www.jungundnaiv.de/},
  year = {undated},
  note={Accessed 25.06.2025}
}

@inproceedings{Klie_Bugert_Boullosa_EckartdeCastilho_Gurevych_2018,
  address={Santa Fe, New Mexico, USA},
  title={The INCEpTION Platform: Machine-Assisted and Knowledge-Oriented Interactive Annotation},
  booktitle={Proceedings of System Demonstrations of the 27th International Conference on Computational Linguistics (COLING 2018)},
  author={Klie, Jan-Christoph and Bugert, Michael and Boullosa, Beto and Eckart de Castilho, Richard and Gurevych, Iryna},
  year={2018}
}

@book{Nordin_Schiappa_2024, 
address={New York}, 
edition={2}, 
title={Argumentation: Keeping Faith with Reason},
ISBN={978-1-003-41526-8}, url={https://www.taylorfrancis.com/books/9781003415268}, DOI={10.4324/9781003415268}, publisher={Routledge}, author={Nordin, John P. and Schiappa, Edward}, year={2024}, month=may, language={en} }

@book{Toulmin_2003, place={Cambridge}, edition={2}, title={The Uses of Argument}, publisher={Cambridge University Press}, author={Toulmin, Stephen E.}, year={2003}}

@book{van_Dijk_2000, address={London}, title={Ideology: A Multidisciplinary Approach}, url={https://sk.sagepub.com/dict/mono/ideology/toc}, DOI={10.4135/9781446217856}, publisher={SAGE Publications Ltd}, author={van Dijk, Teun}, year={2000} }

@article{hayek_2024,
	title = {Media Framing of Government Crisis Communication During {Covid-19}},
	volume = {12},
	rights = {https://creativecommons.org/licenses/by/4.0},
	issn = {2183-2439},
	url = {https://www.cogitatiopress.com/mediaandcommunication/article/view/7774},
	doi = {10.17645/mac.7774},
	pages = {7774},
	journal = {Media and Communication},
	author = {Hayek, Lore},
	urldate = {2026-07-13},
	date = {2024-04-30},
    year = {2024}
}

@inproceedings{blokker-etal-2020-swimming,
    title = "Swimming with the Tide? Positional Claim Detection across Political Text Types",
    author = "Blokker, Nico  and
      Dayanik, Erenay  and
      Lapesa, Gabriella  and
      Pad{\'o}, Sebastian",
    editor = "Bamman, David  and
      Hovy, Dirk  and
      Jurgens, David  and
      O'Connor, Brendan  and
      Volkova, Svitlana",
    booktitle = "Proceedings of the Fourth Workshop on Natural Language Processing and Computational Social Science",
    month = nov,
    year = "2020",
    address = "Online",
    publisher = "Association for Computational Linguistics",
    url = "https://aclanthology.org/2020.nlpcss-1.3/",
    doi = "10.18653/v1/2020.nlpcss-1.3",
    pages = "24--34"
}
